\documentclass[10pt,journal]{IEEEtranTCOM}

\usepackage[english]{babel}
\usepackage[usenames]{color}
\usepackage[cp1250]{inputenc}
\usepackage{amsfonts}
\usepackage{amsthm}
\usepackage{graphicx}
\usepackage{epsfig}
\usepackage{mathrsfs}
\usepackage{amsmath}
\usepackage{algorithm}
\usepackage{algorithmic}

\theoremstyle{plain}

\begin{document}
\newcommand{\bea}{\begin{eqnarray}}
\newcommand{\eea}{\end{eqnarray}}
\newcommand{\be}{\begin{equation}}
\newcommand{\ee}{\end{equation}}
\newcommand{\beas}{\begin{eqnarray*}}
\newcommand{\eeas}{\end{eqnarray*}}
\newcommand{\bs}{\backslash}
\newcommand{\bc}{\begin{center}}
\newcommand{\ec}{\end{center}}
\def\SC {\mathscr{C}}

\title{Exploiting statistical dependencies of time series\\ with hierarchical correlation reconstruction }
\author{\IEEEauthorblockN{Jarek Duda}\\
\IEEEauthorblockA{Jagiellonian University,
Golebia 24, 31-007 Krakow, Poland,
Email: \emph{dudajar@gmail.com}}}
\maketitle

\begin{abstract}
While we are usually focused on forecasting future values of time series, it is often valuable to additionally predict their entire probability distributions, e.g. to evaluate risk, Monte Carlo simulations. On example of time series of $\approx$ 30000 Dow Jones Industrial Averages, there will be presented application of hierarchical correlation reconstruction for this purpose: MSE estimating polynomial as joint density for (current value, context), where context is for example a few previous values. Then substituting the currently observed context and normalizing density to 1, we get predicted probability distribution for the current value.
In contrast to standard machine learning approaches like neural networks, optimal polynomial coefficients here have inexpensive direct formula, have controllable accuracy, are unique and independently calculated, each has a specific cumulant-like interpretation, and such approximation can asymptotically approach complete description of any real joint distribution - providing universal tool to quantitatively describe and exploit statistical dependencies in time series, systematically enhancing ARMA/ARCH-like approaches, also based on different distributions than Gaussian which turns out improper for daily log returns. There is also discussed application for non-stationary time series like calculating linear time trend, or adapting coefficients to local statistical behavior.

\end{abstract}
\section{Introduction}

\begin{figure}[t!]
    \centering
        \includegraphics{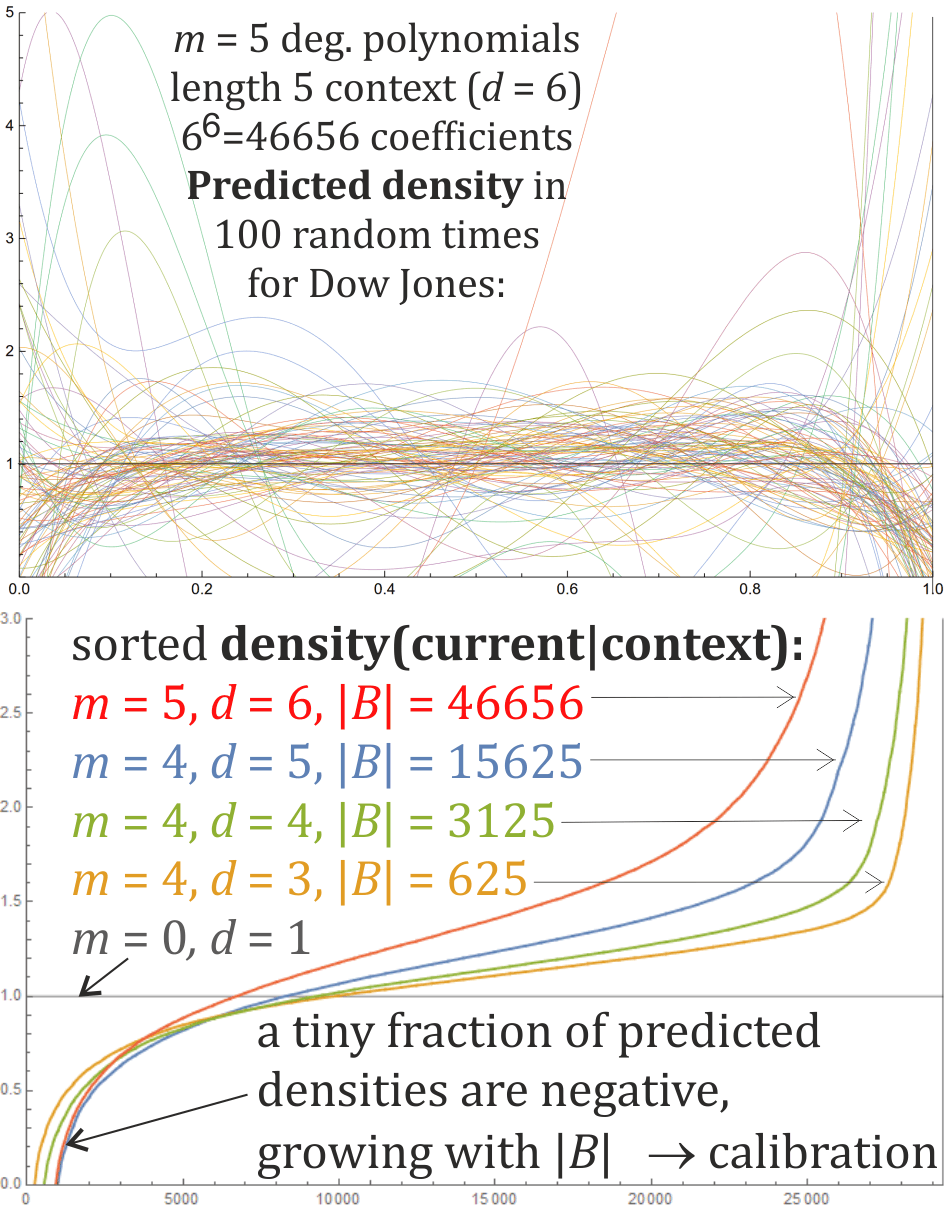}
        \caption{Top: degree $m=5$ polynomials (integrating to 1) on $[0,1]$ range predicting probability density based on length 5 context (previous 5 values) in 100 random positions of analysed sequence (normalized Dow Jones Industrial Averages): joint density for $d=1+5=6$ variables (current value and context) was MSE fitted as polynomial, then substituting the current context and normalizing to integrate to 1, we get predicted density for the current value. We can see that some predicted polynomials go below 0, what can be calibrated to be interpreted as low positive (Fig. \ref{calib}, \ref{plots}). Predicted densities are usually close to  marked $\rho=1$ uniform density (obtained if not using context), but often localize improving prediction. Bottom: evaluation plot by sorting predicted densities for the actual current values in all 29349 situations: in $\approx 20\%$ of cases it gives worse prediction than $\rho=1$  (without using context), but in the remaining cases it is essentially better. The number of coefficients in the used basis is $|B| = (m+1)^d$. We can see that prediction generally improves (higher density) with growing number of coefficients, however, beside growing computational cost, it comes with overfitting (leading e.g. to negative densities) - polynomial approaches sum of Dirac deltas in points of the sample .}
       \label{summ}
\end{figure}
Modeling spatial or temporal statistical dependencies between observed values is a difficult task required in many applications. Standard approaches like correlation matrix, PCA (principal component analysis) approximate this behavior with multivariate gaussian distribution. Further corrections can be extracted by approaches like GMM (gaussian mixture model), KDE (kernel density estimation)~\cite{ker1} or ICA (independent component analysis)~\cite{ica}, but they have many weaknesses like lack of error control, large freedom of parameters and varying their number, or focusing on a specific types of distributions.

Approximating observed data sample with a polynomial is universal approach used in many fields of science, can provide as close description as needed. It turns out also very advantageous for density estimation, including  multivariate joint distribution (\cite{me1,me2}), especially if variables are normalized to approximately uniform distribution on $[0,1]$ with CDF of approximated distribution like in copula theory~\cite{copula}, to properly model tails, optimize weights and standardize coefficients.

Using orthonormal basis $\rho(x)=\sum_f a_f f(x)$, it turns out that mean-square (MSE, $L^2$) optimization leads to estimated coefficients being just averages over the observed sample: $a_f = \frac{1}{n} \sum_{i=1}^n f(x^i)$. For multiple variables we can use basis of products of 1D orthornormal polynomials. On example of DJIA time series \footnote{Source of DJIA time series: http://www.idvbook.com/teaching-aid/data-sets/the-dow-jones-industrial-average-data-set/}, with results summarized in Fig. \ref{summ}, it will be used for prediction of current probability distribution based on a few previous values.

Financial time series are usually modelled with ARMA-ARCH-like simple models~\cite{arma} for this purpose, exploiting pairwise dependencies between first two moments. Presented approach allows to systematically expand it in MSE-optimal way for including any moments among any number of variables, also for non-stationary time series. Additionally, standard models are usually based on Gaussian distribution, which turns out improper e.g. for daily log returns as we can see in a few figures here (especially Fig. \ref{eval}) - presented approach can be based on any distribution (e.g. Laplace), using it for normalization of variables.

Finally, asymptotically (with basis and sample size $\to\infty$) we can approach complete description of any statistical dependencies - any real joint distribution of observed variables. Coefficients can be cheaply calculated as just averages, are unique and independently calculated, for stationary time series we can control their accuracy. Each has also a specific interpretation: resembling cumulants, but being much more convenient for reconstructing probability distribution - instead of the difficult problem of moments~\cite{mom}, here they are just coefficients of polynomial as density. Inconvenience of parameterizing density with polynomial are occurring negative predictions, which can be interpreted as small positive - we can optimize this calibration based on the data sample as discussed later (Fig. \ref{calib}, \ref{plots}).

\begin{figure}[t!]
    \centering
        \includegraphics{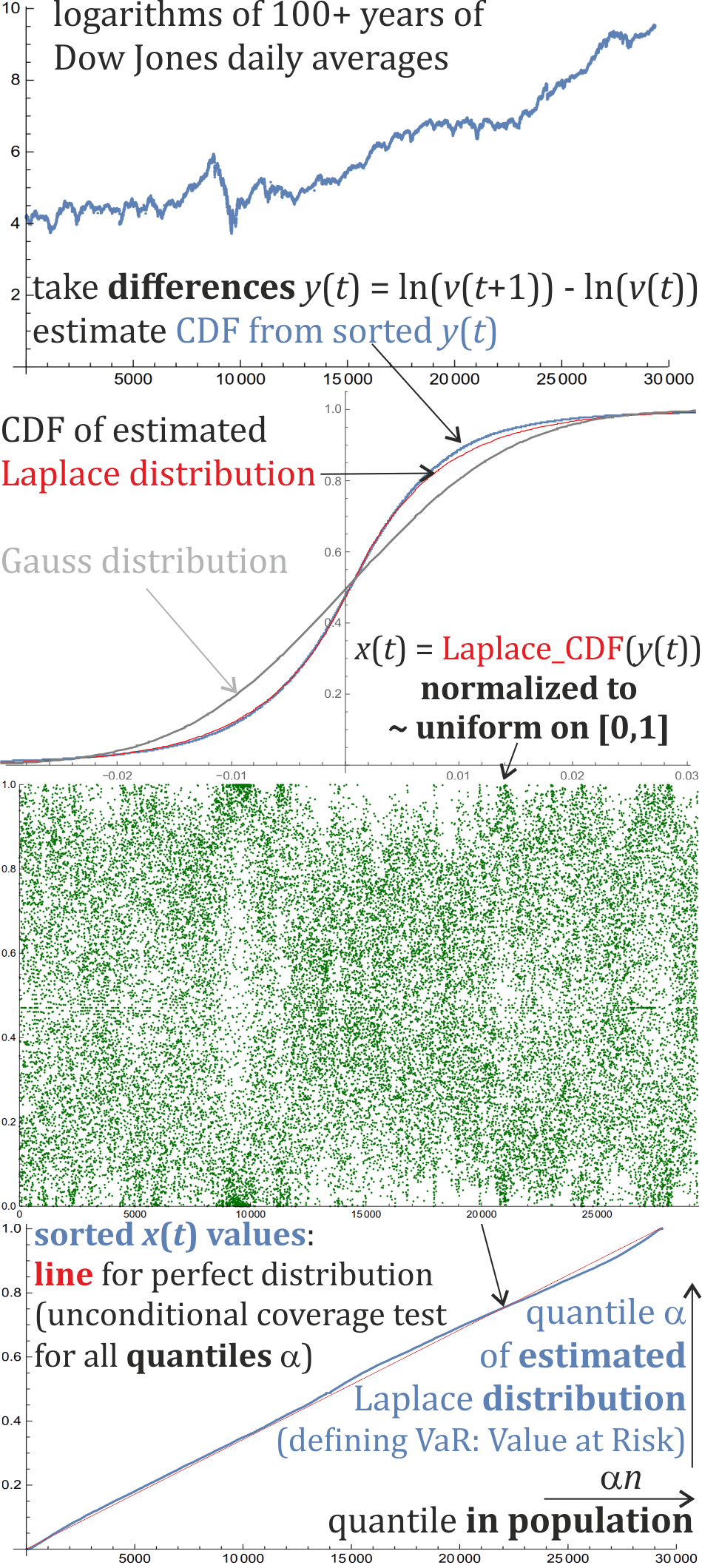}
        \caption{Normalization of the original variable to nearly uniform on $[0,1]$ (marked green) used for further correlation modelling. The original sequence $\{v^t\}$ of 29355 Dow Jones daily averages (over 100 years) is first logarithmized (top plot), then we take differences $y^t=\ln(v^{t+1})-\ln(v^t)$. Sorting $\{y^t\}$ we get its approximated CDF, which, in contrast to standard Gaussian assumption, turns out in good agreement with Laplace distribution ($\mu\approx 0.00044,\ b\approx 0.0072$) - estimated and drawn (red) in the second plot. The marked green next plot is the final $x^t=\textrm{CDF}_{Laplace(\mu,b)}(y^t)$ sequence used for further correlation modeling. The bottom plot shows sorted $\{x^t\}$ values to verify that they come from nearly uniform distribution (line) - its inaccuracy will be repaired later with fitting polynomial (Fig. \ref{polyn}). It corresponds to standard unconditional coverage test~\cite{risk} checking if modelled quantiles agree with population percentages - it is usually done only for a few extreme quantiles (e.g. 1\%, 5\%), getting line for such sorted CDF$(y)$ would mean perfect agreement for all quantiles. }
       \label{norm}
\end{figure}

In the discussed example: analysis of DJIA time series, we will first normalize the variables to nearly uniform probability distribution on $[0,1]$: by considering differences of logarithms (so called log returns), and transforming them by CDF (cumulative distribution function) of approximated distribution (usually Laplace) as shown in Fig. \ref{norm} (and \ref{cor}). In copula theory~\cite{copula} such normalization is made with empirical distribution, then there is used usually single parameter distribution to describe dependencies. Here we normalize with parametric distribution instead, repairing inaccuracy by fitting polynomial, described by as large number of parameters as needed - it allows for  flexibility, including modelling evolution of polynomial coefficients for non-stationary time series. Additionally, thanks to normalization, coordinates correspond to quantiles e.g. 1/2 to median, length to percentage of population - providing proper weights for optimized $L_2$ distance.

After normalization $x=\textrm{CDF}(y)$, looking at $d$ successive positions of $x$, if uncorrelated they would come from $\rho\approx 1$ distribution on $[0,1]^d$. Its corrections as linear combination of orthonormal basis of polynomials can be inexpensively and independently calculated, providing unique and asymptotically complete description of statistical dependencies between these neighboring values. Treating $d-1$ of them as earlier context, substituting their values and normalizing to 1, we get predictions of (conditional) probability distribution for the current value as summarized in Fig. \ref{summ}.

Handling of non-stationary time series is also proposed: by replacing $a_f = \frac{1}{n} \sum_{i=1}^n f(x^i)$ global average with local averages over past values with exponentially decaying weights, or using interpolation with time as additional dimension.

Presented approach can be naturally extended to multivariate time series, e.g. stock prices of separate companies to model their statistical dependencies, what is presented in \cite{me4} on example of yield curve parameters, here there will be presented example of modelling various statistical dependencies of 29 of Dow Jones companies.

More relaxed treatment of values of such polynomials as cumulant-like features and contributions, also combining with discrete variables, can be found in \cite{me5} for credibility evaluation of income data.
\section{Normalization to nearly uniform density}
We will first work on example of Dow Jones Industrial Averages time series $\{v^t\}_{t=1..n_0}$ for $n_0=29355$. As financial data usually evolve in multiplicative not additive manner, we will work with $\ln(v^t)$ to make it additive.

Time series are usually normalized to allow assumption of stationary process: such that joint probability distribution does not change when position is shifted. The standard approach, especially for gaussian distribution, is to subtract mean value, then divide by the standard deviation.

However, above normalization does not exploit local dependencies between values, what we are interested in. Using experience from data compression (especially lossless image e.g. JPEG LS~\cite{loco}), we can use a predictor for the next value based on its local context: for example a few previous values (2D neighbors for image compression), or some more complex features (e.g. using averages over time windows, or dimensionality reduction methods like PCA), then model probability distribution of difference from the predicted value (residue).

Considering simple linear predictors: $v^t\approx \beta_0+\sum_{i=1}^k \beta_i v^{t-i}$ as in ARMA-like models, we can use optimize  $\{b_k\}$ parameters to minimize mean square error. For 2D image such optimization leads to approximate parameters $v_{x,y}\approx 0.8 v_{x-1,y}- 0.3 v_{x-1,y-1}+ 0.2 v_{x,y-1}+0.3 v_{x+1,y-1}$. For Dow Jones sequence such optimization leads to nearly negligible weights for all but the previous value. Hence, for simplicity we will just operate on logarithmic returns:
\be y^t=\ln(v^{t+1})-\ln(v^{t})\ee time series, where the number of possible indexes has been reduced by 1 due to shift: $n_1=n_0-1$.

Such sequences of differences from predictions (residues) are well known in data compression to have nearly Laplace distribution - density:
\be g(y)=\frac{1}{2b}\exp\left(-\frac{|y-\mu|}{b}\right) \ee
where maximum likelihood estimation of parameters is just: $\mu=$ median of $y$, $b=$ mean of $|y-\mu|$. We can see in Fig. \ref{norm} that CDF from sorted $y_t$ values has decent agreement with CDF of Laplace distribution. Otherwise, there can be used e.g. generalized normal distribution~\cite{gengaus}, also called exponential power distribution (EPD) or generalized error distributions, which includes both gaussian and Laplace distribution - its optimization is presented in Fig. \ref{eval}. Student's t or stable distributions (Levy)~\cite{stable} might be also worth considering as they include heavy tail distributions. These distributions have also known asymmetric variants - which can be considered if two directions have essentially different tails. 

We could also use a varying distribution for normalization $y^t=\textrm{CDF}_t(x^t)$, using for example CDF of Gaussian of width depending on previous value using coefficients from ARCH model. This way we could directly enhance some successful standard model - by using it for normalization, than fitting polynomial to exploit complex further statistical dependencies: missed by the low parameter model.

\begin{figure}[b!]
    \centering
        \includegraphics{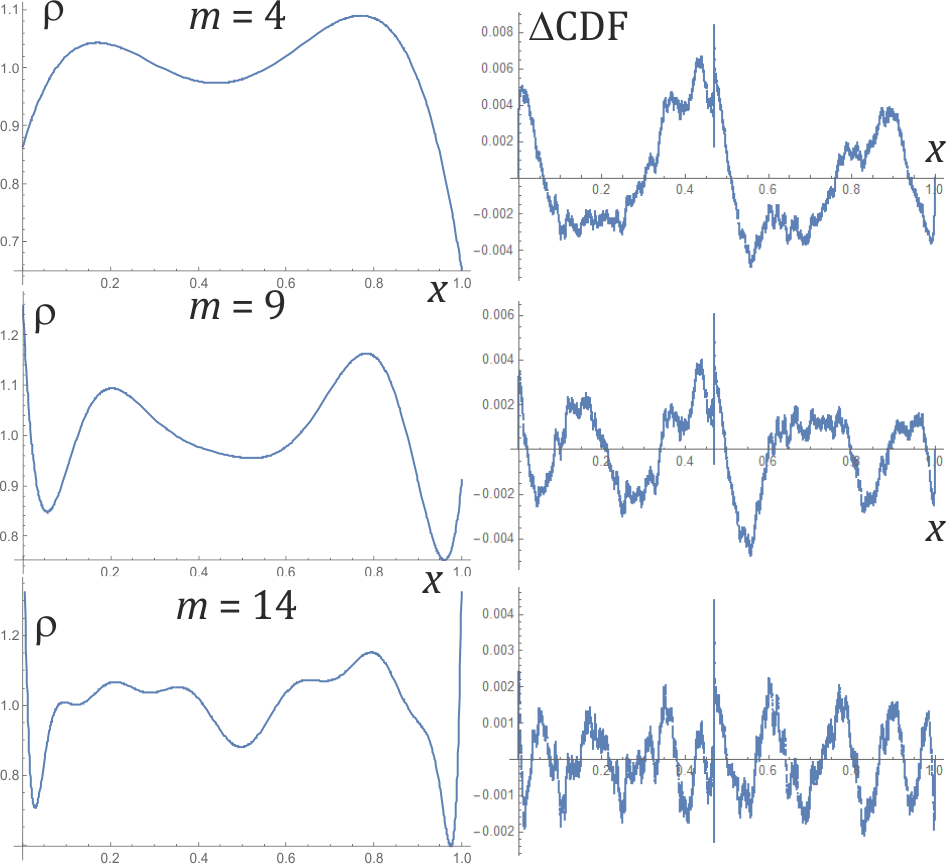}
        \caption{Modelling probability distribution as independent variable $(d=1)$ using degree $m$ polynomials: $\rho(x)=\sum_{j=0}^m a_j f_j(x)$. After normalization with CDF of Laplace distribution, we should have $\rho\approx 1$. Here we repair its inaccuracy with estimated polynomial (left column), corresponding to the plot in the bottom of Fig. \ref{norm} - the right column contains differences between empirical CDF and such fitted polynomial. Obviously this difference reduces with degree $m$, however, we can see that it contains a growing number $(\approx m)$ of oscillations (Runge's phenomenon). }
       \label{polyn}
\end{figure}

For simplicity there is mainly used Laplace distribution here to normalize our variables to nearly uniform in $[0,1]$, what allows to compactify the tails, improve performance (weight proportional to population) and normalize further coefficients:
\be x^t=G(y^t)\qquad\textrm{where}\quad G(y)=\int_{-\infty}^y g(y')\, dy' \label{trans}\ee
is CDF of used distribution (Laplace here). We can see in Fig. \ref{norm} (and later \ref{cor}) that this final $x_t$ sequence has nearly uniform probability distribution. Its corrections will be included in further estimation of polynomial as (joint) probability distribution, like presented later in Fig. \ref{polyn}.

We will search for $\rho_X (x)$ density. To remove transformation (\ref{trans}) to get final $\rho_Y(y)$ density, observe that $P(y'=G^{-1}(x) \leq y)=P(x \leq G(y))$. Differentiating over $y$, we get
\be \rho_Y (y) = \rho_X(G(y))\cdot g(y).\label{densn}\ee
\section{Hierarchical correlation reconstruction}
After normalization we have $\{x^t\}$ sequence with nearly uniform density, marked green in Fig. \ref{norm} here. Taking its $d$ succeeding values, if uncorrelated they would come from nearly uniform distribution on $[0,1]^d$ - difference from uniform distribution describes statistical dependencies in our time series. We will use polynomial to describe them: estimate joint density for $d$ succeeding values of $x$.

Define $x_i^t=x^{t-i+1}$ for $i=1,\ldots,d$ and $t=1,\ldots,n$, $n=n_1-d+1$. They form $\textbf{x}^t=\{x^t_i\}_{i=1..d}\in [0,1]^d$ vectors containing value with its context - we will model (joint) probability density of these vectors as polynomial. Generally we can also use more sophisticated contexts, for example average of a few earlier values (e.g. $(x_{t-5}+x_{t-6})/2$) as a single context variable to include longer range correlations, or e.g. some macroeconomical variables to exploit also dependence from additional information. Dimensionality reduction techniques like PCA can be used to include dependence from a large number of variables. Normalization to nearly uniform density is recommended for the predicted values ($x^t_1$), for context values it might be better to omit it, especially when absolute values are important like for image compression.

Finally assume we have $\{\textbf{x}^t\}_{t=1,\ldots,n}\subset[0,1]^d$ vector sequence of value with its context, we would like to model density of such vectors as polynomial. It turns out~\cite{me1} that using orthonormal basis, which for multidimensional case can be products of 1D orthonormal polynomials, mean square ($L^2$) optimization leads to extremely simple formula for estimated coefficients:
$$ \rho(\textbf{x})=\sum_{\textbf{j}\in\{0\ldots m\}^d}a_\textbf{j} f_\textbf{j}(\textbf{x})=\sum_{j_1\ldots j_d=0}^m a_\textbf{j}\,f_{j_1}(x_1)\cdot \ldots\cdot f_{j_d}(x_d) $$
\be \textrm{with estimated coefficients:}\quad a_\textbf{j} = \frac{1}{n} \sum_{t=1}^n f_\textbf{j}(\textbf{x}^t) \ee

The basis used this way has $|B|=(m+1)^d$ functions, generally it seems worth to consider different $m_i$ for separate coordinates $(|B|=\prod_{i=1}^d (m_i+1))$, or restrict to include correlations only from e.g. pairs by using $\textbf{j}$ with at most two nonzero indexes. Beside inexpensive calculation, this simple approach has also very convenient property of coefficients being independently calculated, giving each $\textbf{j}$ unique value and interpretation, and controllable error. Independence also allows for flexibility of considered basis - instead of using all $\textbf{j}$, we can focus on more promising ones: with larger absolute value of coefficient, replacing negligible $a_\textbf{j}$. Instead of mean square optimization, we can use often preferred: likelihood maximization~\cite{me2}, but it requires additional iterative optimization and introduces dependencies between coefficients.

Above $f_j$ 1D polynomials are orthonormal in $[0,1]$: $\int_0^1 f_j(x) f_{k}(x) dx = \delta_{jk}$, getting (rescaled Legendre): $f_0=1$ and for $j=1,2,3,4,5$ correspondingly:
$$\sqrt{3}(2x-1), \sqrt{5}(6x^2-6x+1), \sqrt{7}(20x^3-30x^2+12x-1),$$
$$3(70x^4-140x^3+90x^2-20x+1), $$
$$\sqrt{11}(252x^5-630x^4+560x^3-210x^2+30x-1).$$

\begin{figure}[t!]
    \centering
        \includegraphics{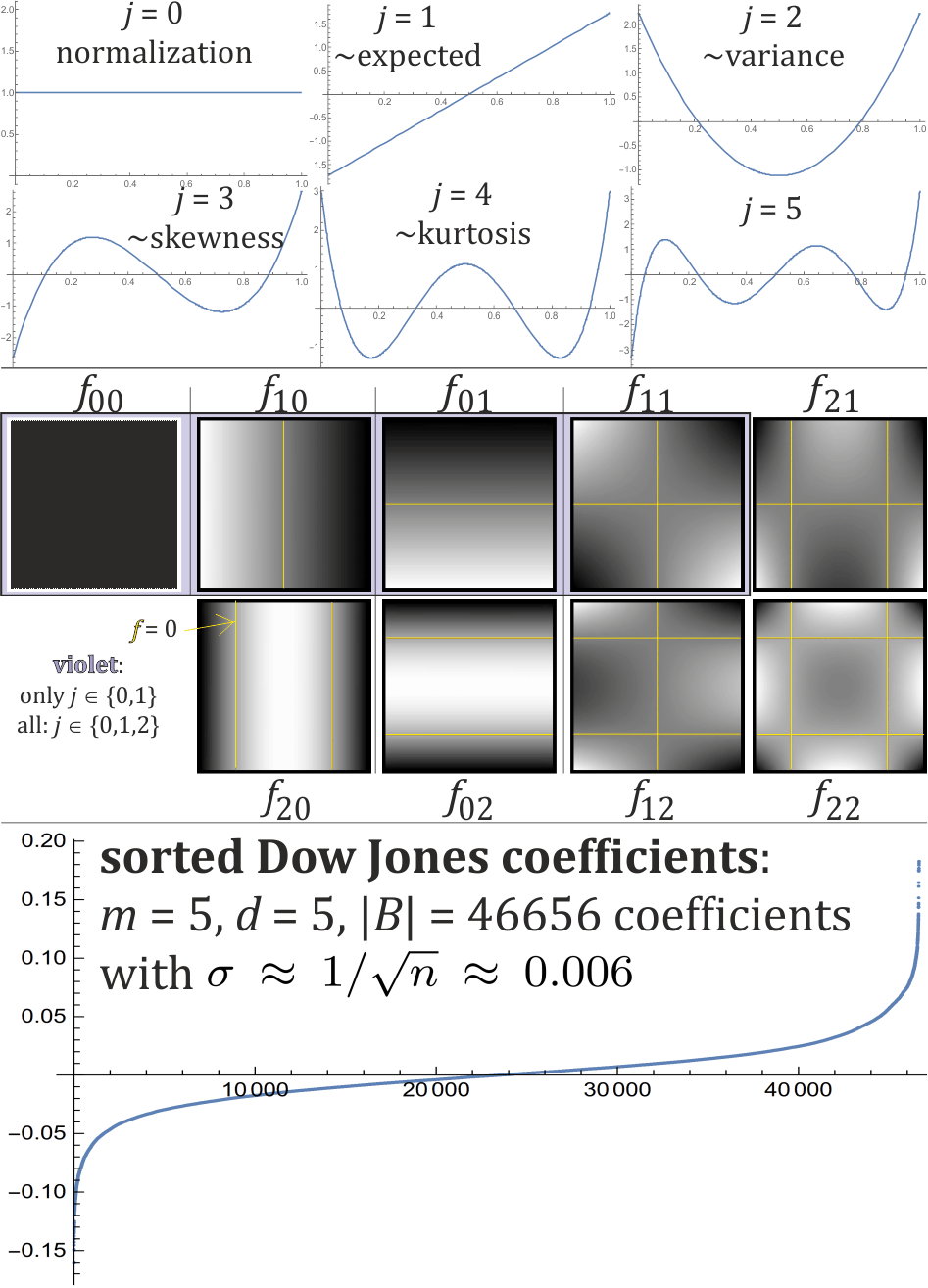}
        \caption{Top: the first 6 of used 1D orthonormal basis of polynomials $(\langle f,g\rangle=\int_0^1 f g\, dx)$: $j=0$ coefficient guards normalization, the remaining functions integrate to 0, and their coefficients describe perturbation from uniform distribution. These coefficients have similar interpretation as cumulants, but are more convenient for density reconstruction. Center: 2D product basis for $j\in \{0,1,2\}$. The $j=0$ coordinates do not change with corresponding perturbation. Bottom: sorted calculated coefficients (without $a_{000000}=1$) for DJIA sequence, $m=5$ and length 5 context ($d=6$) modelling. Assuming stationarity, for uniform distribution their standard deviation would be $\sigma\approx 1/\sqrt{n}\approx 0.006$, exceeded here more than tenfold by many coefficients - allowing to conclude that they are essential: not just a noise.}
       \label{coef}
\end{figure}

Their plots are in the top of Fig. \ref{coef}. $f_0$ corresponds to normalization. The $j=1$ coefficient decides about reducing or increasing the mean - has similar interpretation as expected value. Analogously $j=2$ coefficient decides about focusing or spreading given variable, similarly as variance. And so on: further $a_j$ has similar interpretation as $j$-th cumulant, however, while reconstructing density from moments is a difficult problem, presented description is directly coefficients of polynomial estimating the density.

For multiple variables, $a_\textbf{j}$ describes only correlations between $C=\{i:j_i>0\}$ coordinates, does not affect $j_i=0$ coordinates, as we can see in the center of Fig. \ref{coef}. Each such mixed coefficient has also a specific interpretations here, for example $a_{11}$ decides between increase and decrease of second variable with increase of the first, $a_{12}$ analogously decides focus or spread of the second variable.

Assuming stationary time series (fixed joint distribution of $\textbf{x}^t\in [0,1]^d$), errors of such estimated coefficients come from approximately gaussian distribution:
\be \tilde{a}-a \sim \mathcal{N}\left(0,\frac{1}{\sqrt{n}}\sqrt{\int (f_j-a_j)^2 \rho\, d\textbf{x}}\right) \label{error}\ee
For $\rho=1$ the integral has value 1, getting $\sigma=1/\sqrt{n}\approx 0.006$ in our case. As we can see in bottom of Fig. \ref{coef}, a few percents of coefficients here are more that tenfold larger: can be considered as essential, not a result of noise.

Here is a list of the largest $|a_\textbf{j}|>0.14$ coefficients for Dow Jones normalized series (beside $a_{000000}=1$) in $d=6$, $m=5$ case. It neglects shifted sequences, for example $a_{200200}\approx a_{020020} \approx a_{002002}$.

\noindent Positive:

 $ a_{200200}\approx 0.184867 \qquad  a_{200002}\approx 0.183297 $

 $a_{200020}\approx 0.178384 \qquad  a_{202000}\approx 0.177606  $

 $a_{554555}\approx 0.176333 \qquad  a_{220000}\approx 0.176184  $

 $a_{554535}\approx 0.169778 \qquad a_{554355}\approx 0.161684  $

 $a_{545445}\approx 0.156764 \qquad a_{555555}\approx 0.149727 $

$ a_{555355}\approx 0.147934 \qquad  a_{454523}\approx 0.145962 $

\noindent Negative:

$a_{555552}\approx-0.170723 \qquad a_{344544}\approx-0.166773$

$a_{455235}\approx-0.156860 \qquad a_{342544}\approx-0.149314$

$a_{455255}\approx-0.147201 \qquad a_{555451}\approx-0.146523$

$a_{555532}\approx-0.145356 \qquad a_{553451}\approx-0.143087$

$a_{555352}\approx-0.142076 \qquad a_{355451}\approx-0.140343$

\begin{figure}[b!]
    \centering
        \includegraphics{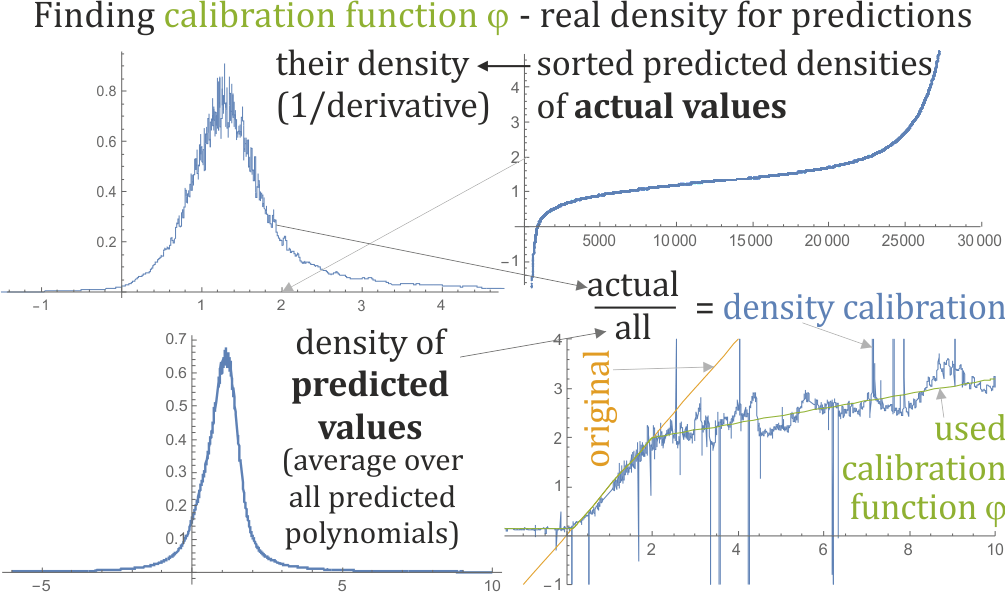}
        \caption{Finding calibration function $\varphi$: proper interpretation of predicted densities $(\rho\to\varphi(\rho))$, including negative as small positive. Fig. \ref{plots} shows example of its use. Top right: sorted predicted densities for actually observed values: evaluation plot from Fig. \ref{summ} for $m=5, d=6$ case. Top left: density of its values, calculated numerically as 1/derivative from top right plot, it integrates to $\approx 1$. Bottom left: density of all predicted values from all $n$ predicted polynomials - it can be found analytically, but it is simpler to do it numerically: $[0,1]$ range was approximated with size 1000 regular lattice, getting $1000n$ values from all polynomials. They were sorted getting empirical distribution function, which 1/derivative is the presented density, integrating to $\approx 1$. Bottom right: finding calibration function from dividing both densities - such that we should interpret predicted polynomial $\rho$ as density $\varphi(\rho)$. Orange line shows using $\varphi(z)=z$ in this place, and obviously has a problem for predicted densities below zero. It agrees with blue plot trend on $[0,2]$ range, confirming that predictions in this range can be interpreted as practically unchanged. For application we should smoothen and parameterize this function, like using the green plot:  $\varphi(z)=\max(0.15,\min(z,0.15z+1.7))$ including small positive density for values below 0 and slower positive trend above 2. Alternatively, calibration function can be chosen by assuming some parametrization like $\varphi(\rho)=\max(\rho,a)$ for $a\approx 0.2$ and optimizing its parameters to maximize log-likelihood of prediction like in Fig. \ref{eval}.}
       \label{calib}
\end{figure}
\begin{figure}[b!]
    \centering
        \includegraphics{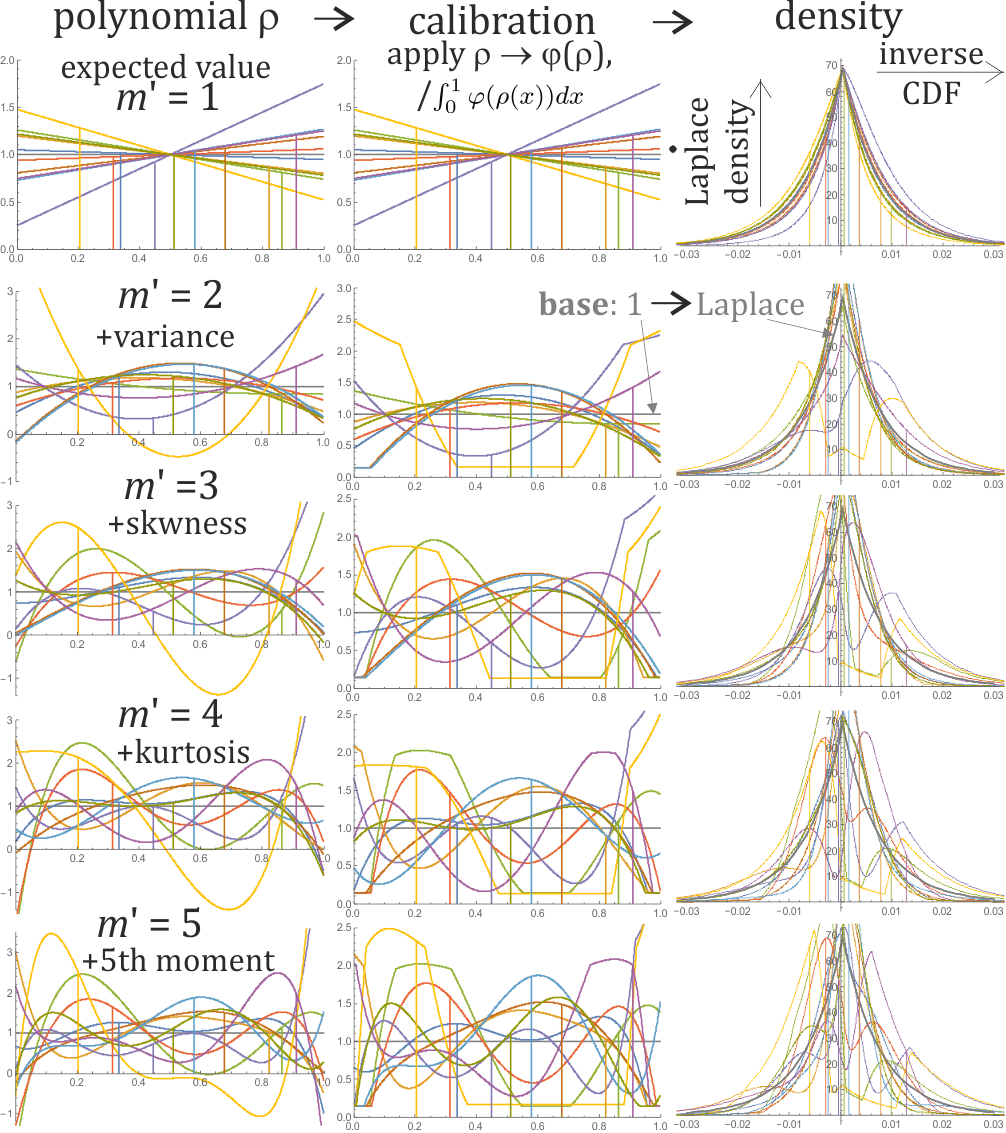}
        \caption{Examples of predicted polynomials for the discussed $m=5, d=6$ DJIA case, their calibration to nonnegative densities, and translating into final densities (removing normalization with Laplace) - for randomly chosen 10 points (moments of time, the same for all rows). However, each row applies only first $m'=1..5$ of predicted cumulant-like coefficients: presented predicted polynomials are of these degrees. For simplicity each use calibration: $\varphi(z)=\max(0.15,\min(z,0.15z+1.7))$ green plot from Fig. \ref{calib}, calculated for the last row (the remaining have slightly different but similar). Applying $\rho\to\varphi(\rho)$ usually damage probabilistic normalization, hence calibration step also includes division by $\int_0^1 \varphi(\rho(x))dx$. Finally to remove normalization with Laplace distribution here, there is applied Laplace $CDF^{-1}$ on horizontal axis, and multiplication by Laplace distribution density on vertical axis (\ref{densn}).}
       \label{plots}
\end{figure}

Each such unique coefficient describes a specific correction from uniform density: by $a_\textbf{j}\, f_{j_1}(x_1)\cdot \ldots\cdot f_{j_d}(x_d)$. For example we can see large positive coefficients for all pairs of $j=2$, what means upward directed parabola for both variables: quantitatively describes how large change in a given day increases probability of large changes in neighboring days. Further coefficients have more complex interpretations, for example large positive $a_{555555}$ means that 6 large increases in a row are preferred, but 6 large decreases are less likely. In contrast, large negative $a_{555552}$ means that larger change 5 days earlier reduces probability of 5 large increases in a row.

Having such density we can use it to predict probability distribution of the current symbol basing on the context (Fig. \ref{summ}): by substituting context to the polynomial and normalizing the remaining 1D polynomial to integrate to 1. This predicted density as polynomial can sometimes go below zero, what needs a separate interpretation as low positive - we can obtain such proper calibration $\rho\to\varphi(\rho)$ from the data sample, as discussed in Fig. \ref{calib}, \ref{plots}.

\section{Adaptivity for non-stationary time series}

\begin{figure}[b!]
    \centering
        \includegraphics{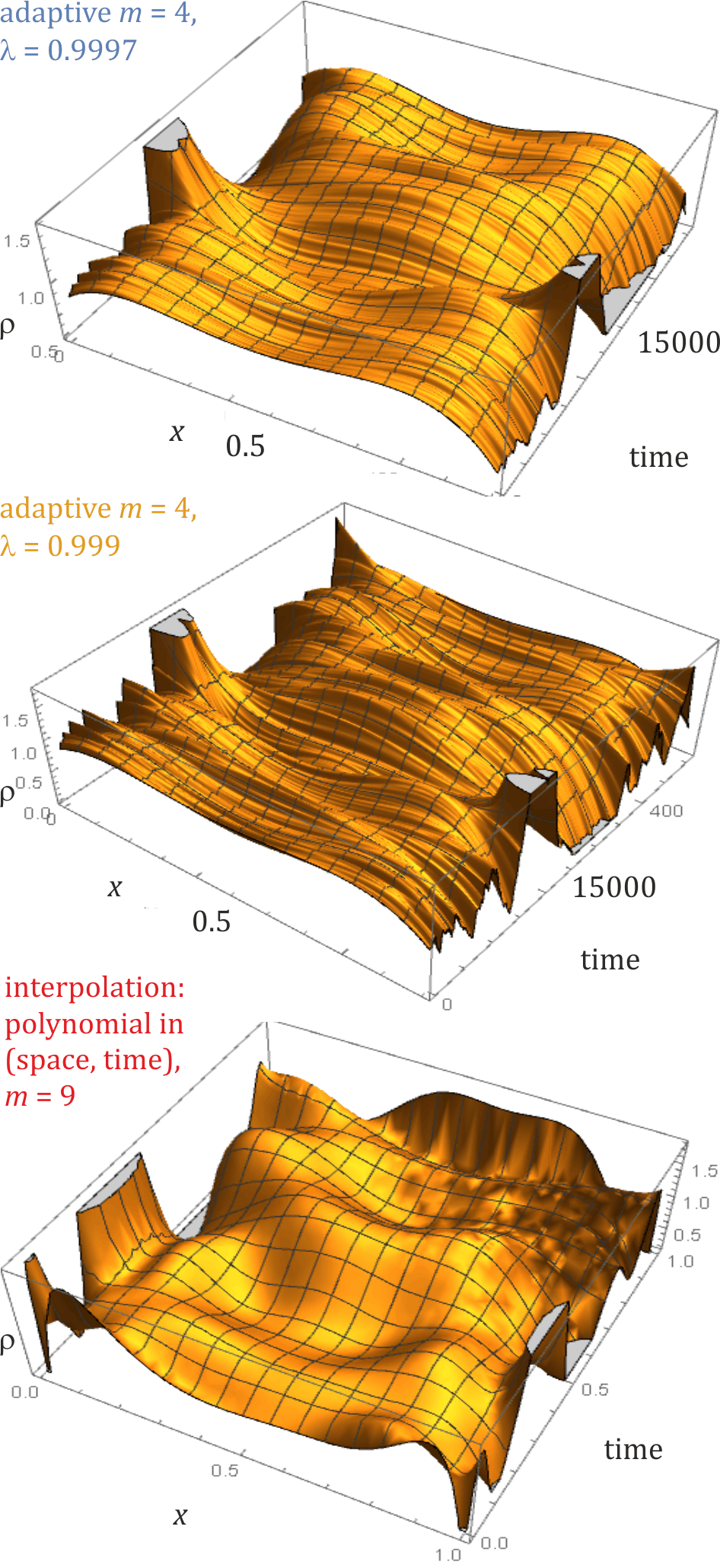}
        \caption{Modelling non-stationary probability distribution of values ($d=1$) - like in Fig. \ref{polyn}, but adapted to inhomogeneous behavior in time. The top two plots used  adaptive averaging $a_f^{t+1} =\lambda a_f^t + (1-\lambda) f(\textbf{x}^t)$ for $m=4$ with two different learning rates $\lambda=0.9997$ or $0.999$. The bottom plot has estimated $m=9$ degree polynomial for density of $(t,x^t)$ variables - in contrast to adaptive averaging, it requires already knowing the future. }
       \label{adaplot}
\end{figure}
\begin{figure}[b!]
    \centering
        \includegraphics{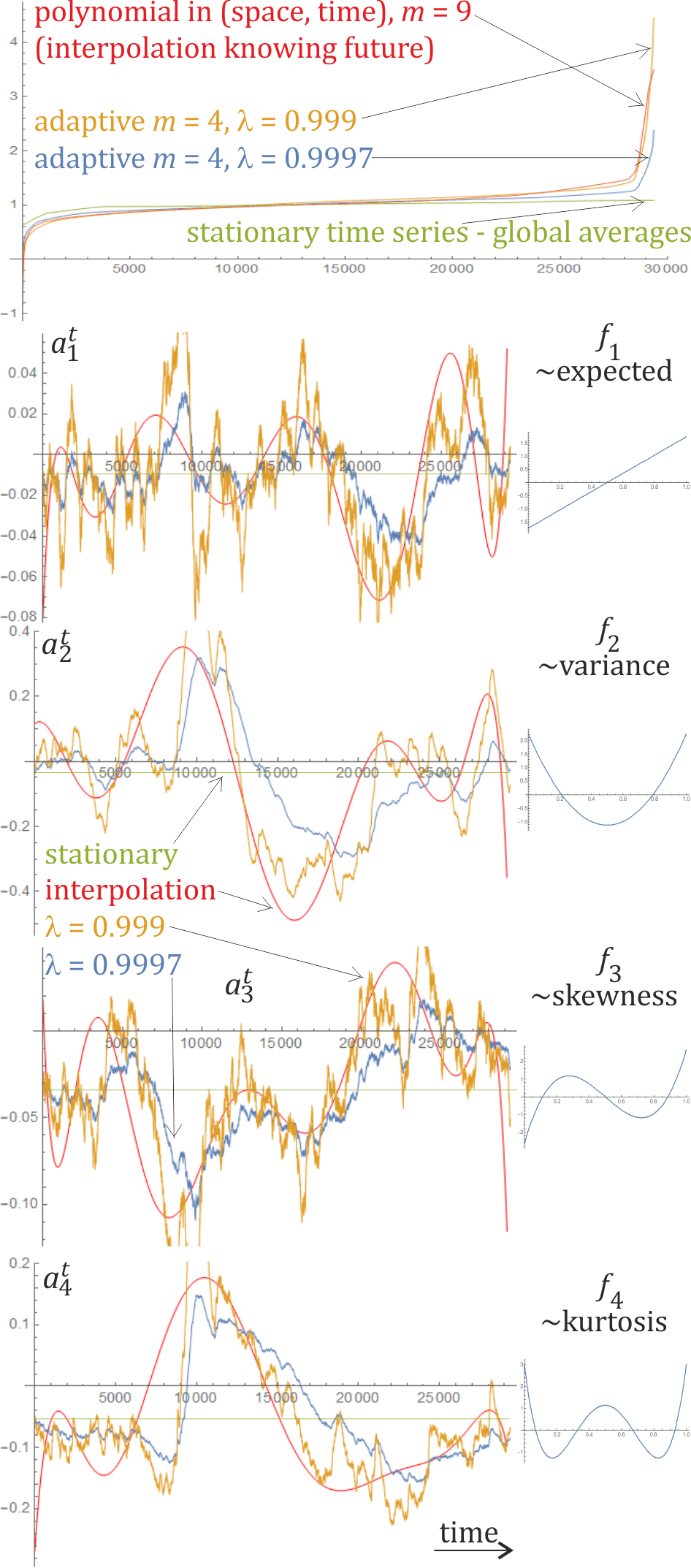}
        \caption{Top: evaluation of results of models presented in Fig. \ref{adaplot} - sorted predicted densities of actual values. They are compared with stationary model: green line, using fixed coefficients being averages over entire time period. Bottom: time dependence of first four coefficients over the time: $\rho^t(x)=\sum_j a_j^t f_j(x)$. They are constant for the stationary model (green lines), degree 9 polynomials for interpolation (red), and noisy curves for adaptive averaging - especially the orange one for relatively low $\lambda=0.999$. The blue curve for $\lambda=0.9997$ is smoother, however, it is at cost of delay (shifted right) - needs more time to adapt to new behavior.        }
       \label{adapt}
\end{figure}
We have previously assumed stationary time series: that joint probability distribution within length $d$ moving time windows is fixed, what is often only approximation for real time series. As coefficients here are just averages over values: $a_f = \frac{1}{n} \sum_{\textbf{x}} f(\textbf{x})$, for coefficients describing local behavior we can use (known in data compression) averaging with exponentially decaying weights~\cite{me2}:
\be a_f^{t+1} =\lambda a_f^t + (1-\lambda) f(\textbf{x}^t)\qquad  \rho^t(x)=\sum_f a_f^t f(x) \label{ada}\ee
for some learning rate $\lambda$: close but smaller than 1 (e.g. $\lambda=0.999$), starting for example with $a_f(0)=0$. Its proper choice is a difficult question: larger $\lambda$ gives smoother behavior, but needs more time to adapt (delay).

For modeling of time trends or a posteriori analysis of historical data (with known future), we can alternatively estimate polynomial for multi-dimensional variable with time as one of coordinates, rescaled to $[0,1]$ range, e.g. $(t/n,\textbf{x}^t)$. This way we estimate behavior of each coefficient as polynomial, allowing e.g. to interpolate to real time, or try to forecast that future trend (as low degree polynomial) will be similar as in the earlier period.

It might be tempting to use this approach also for extrapolation to predict future trends, e.g. rescale time to $[0,1-\epsilon]$ range instead, and look at behavior in time 1. However, such polynomial often has some uncontrollable behavior at the boundaries, suggesting caution while such extrapolation. Other orthonormal families (e.g. sines and cosines) have better boundary behavior - might be more appropriate for such extrapolation, however, discussed earlier modelling of joint distribution with context representing the past is generally a safer approach.

The following three figures present such analysis for discussed DJIA sequence. Figure \ref{polyn} contains estimation of density as polynomial using stationarity assumption - models inaccuracy of Laplace used in normalization. Figure \ref{adaplot} contains its time evolution for non-stationary models: adaptive or interpolation. Figure \ref{adapt} evaluates these approaches and shows time evolution for first 4 cumulant-like coefficients.

\section{Modelling multivariate dependencies}
\begin{figure}[b!]
    \centering
        \includegraphics{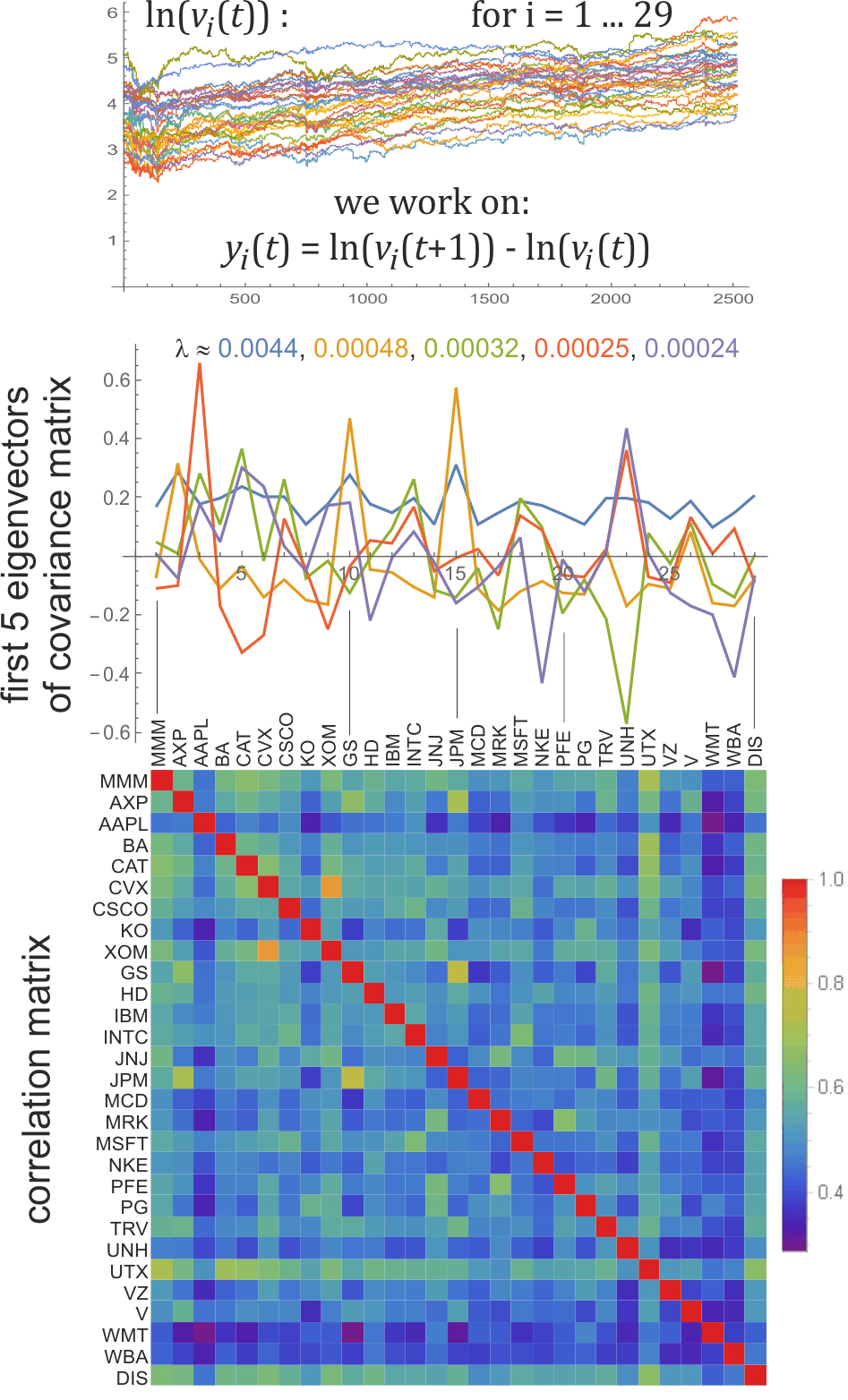}
        \caption{Top: 10-year evolution 2008-2018 of logarithm of stock price of 29 out of 30 Dow Jones companies (without DWDP). We will work on their daily differences: $y_i$. Center: eigenvectors corresponding to 5 largest eigenvalues of their covariance matrix $\left(\Sigma_{ij}=E[(Y_i-E[Y_i])(Y_j-E[Y_j])]\right)$ is a popular tool (PCA) showing directions of largest variance, grouping dependent variables - e.g. financial companies in vector presented as the yellow plot. Bottom: correlation matrix (covariance matrix normalized to unit variance for each variable) - real non-negative, describing strength of dependency between variables. }
       \label{stock}
\end{figure}

Example of predicting probability distribution for multivariate time series is presented in \cite{me4}. To continue the Dow Jones topic, there will be presented application of the discussed methodology to better understand complex statistical dependencies between stock prizes of 30 companies DJIA is calculated from, for example for more accurate wallet analysis. The selection of these companies has changed throughout the history - there will be used the one from September 2018. Daily prices for the last 10 years can be downloaded from NASDAQ webpage (www.nasdaq.com) for all but DowDuPont (DWDP) - there will be used daily close values for 2008-08-14 to 2018-08-14 period ($n_0=2518$ values) for the remaining 29 companies: 3M (MMM), American Express (AXP), Apple (AAPL), Boeing (BA), Caterpillar (CAT), Chevron (CVX), Cisco Systems (CSCO), Coca-Cola (KO), ExxonMobil (XOM), Goldman Sachs (GS), The Home Depot (HD), IBM (IBM), Intel (INTC), Johnson\&Johnson (JNJ), JPMorgan Chase (JPM), McDonald's (MCD), Merck\&Company (MRK), Microsoft (MSFT), Nike (NKE), Pfizer (PFE), Procter\&Gampble (PG), Travelers (TRV), UnitedHealth Group (UNH), United Technologies (UTX), Verizon (VZ), Visa (V), Walmart (WMT), Walgreens Boots Alliance (WBA) and Walt Disney (DIS). We will again work on logarithmic returns.
\begin{figure}[b!]
    \centering
        \includegraphics{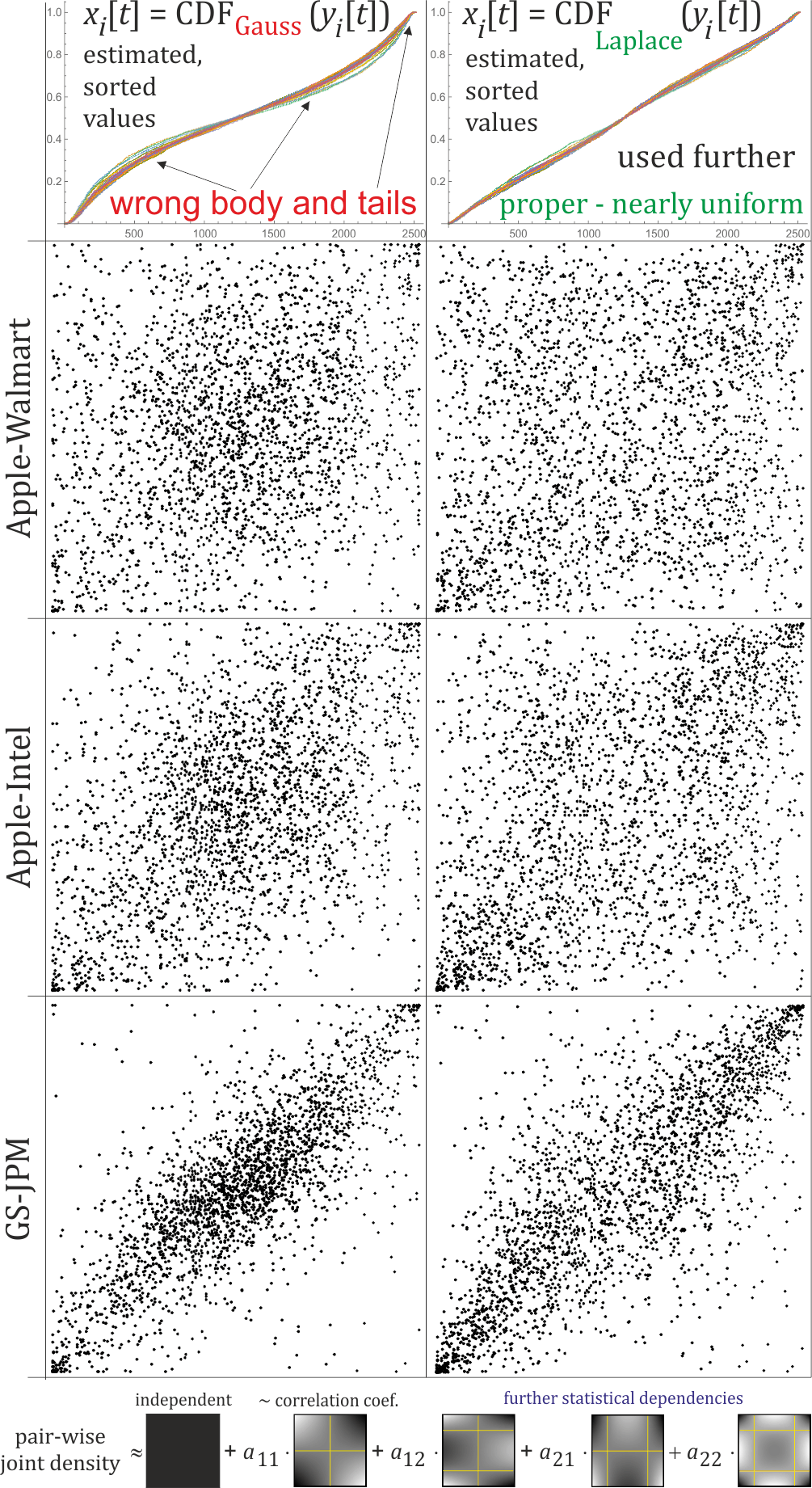}
        \caption{Top: test of normalization (like in Fig. \ref{norm}, unconditional coverage for all quantiles) using Gaussian (left column) or Laplace distribution (right column) for each of 29 variables. Sorting obtained values, we can see that the former is far from uniform distribution (line), especially its tail behavior - what means (often dangerous) underestimation of probability of extreme events. Therefore Laplace distribution is mainly used. Lower diagrams contain dots corresponding to the actual values for three chosen pairs of companies, for nearly uncorrelated Apple-Walmart we can again see that Laplace distribution is more appropriate - leads to distribution closer to uniform. These dots would have uniform distribution on $[0,1]^2$ for uncorrelated variables - to model their statistical dependencies we will MSE fit polynomial to their distribution. Using orthogonal basis presented at the bottom, e.g. coefficient $a_{11}$ corresponds to correlation, what can be seen in similarity of diagrams in Fig. \ref{stock} and \ref{coef1}, $a_{22}$ to variance-variance relation like in ARCH model, and is often significant in economical data (dark edges).}
       \label{cor}
\end{figure}
\begin{figure}[t!]
    \centering
        \includegraphics{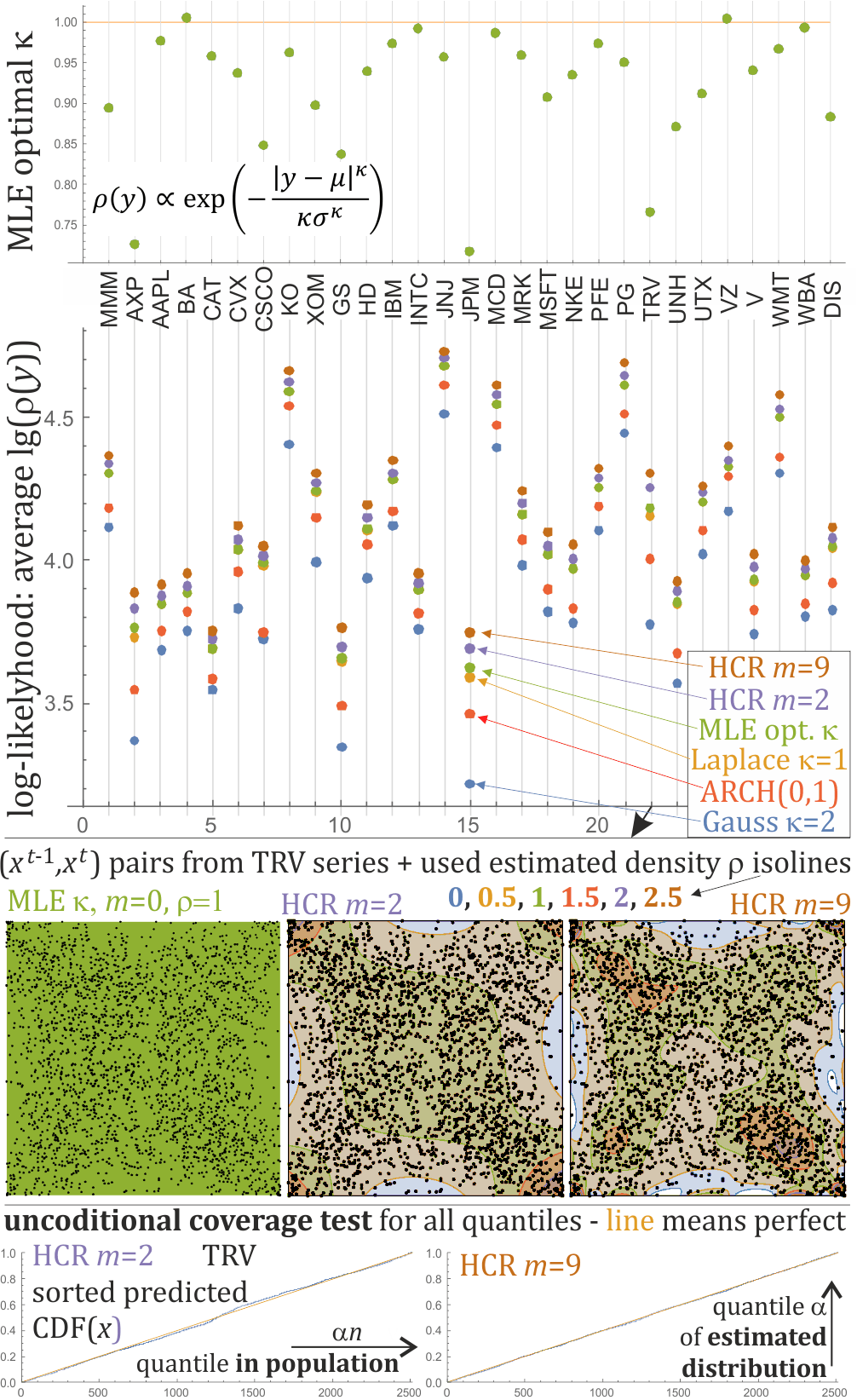}
        \caption{Predictions evaluation for stationary models and at most 1 previous value context for the 29 time series of log returns. Top: $\kappa$ exponents obtained by MLE optimization of exponential power distribution (EPD)~\cite{gengaus} for the 29 series: $\rho(y)\propto \exp(-|y-\mu|^\kappa /\kappa\sigma^\kappa)$, which contains both Gaussian $(\kappa=2)$ and Laplace $(\kappa=1)$ distribution - we can see that most optimal $\kappa$ are even lower than Laplace $\kappa=1$. Second plot: evaluation using log-likelihood: average $\lg(\rho(y))$ of predictions over the series (counted in bits: $\lg\equiv \log_2$). In most cases (beside AXP, JPM, TRV also outlying in top plot), optimal $\kappa$ (green) gives nearly no improvement from Laplace $\kappa=1$ (orange), which is on average 0.21 bits better than Gaussian distribution (blue), what corresponds to $2^{0.21}\approx 1.158$ times larger average density. Improvement of ARMA(0,1) turned out practically negligible (omitted), but ARCH(0,1) is presented (red): choosing variance accordingly to the previous value: $\sigma^t = \alpha_0 + \alpha_1 (y^{t-1})^2$. It is essentially better than pure Gaussian (on average by 0.098 bits), but still far from pure Laplace. The use of optimal $\kappa$ was further improved by modelling $(x^{t-1},x^t)$ joint distribution with polynomial: ARCH-like of degree $m=2$ (violet, 0.035 bits average improvement) and degree $m=9$ (up to 9-th mixed moments, brown, 0.075 bits average improvement from green). Their modeled joint densities are presented in 3 diagrams, for larger $m$ overfitting can happen (not considered), should saturate at theoretical boundary given by some unknown conditional entropy. Bottom: testing unconditional coverage, which is nearly perfect here (line) for all quantiles. The general conclusion is that the first step for improvement is replacing Gaussian with a distribution closer to the real data, e.g. by estimating EPD which contains both Gaussian and Laplace. In case there is some really successful model, we can also enhance it with HCR: use it for normalization $x^t=\textrm{CDF}_t (y^t)$ using its modeled distributions like ARCH Gaussian of varying width, then model e.g. $(x^{t-1},x^t)$ with polynomial to extract and exploit further statistical dependencies from data sample. }
       \label{eval}
\end{figure}

\begin{figure}[b!]
    \centering
        \includegraphics{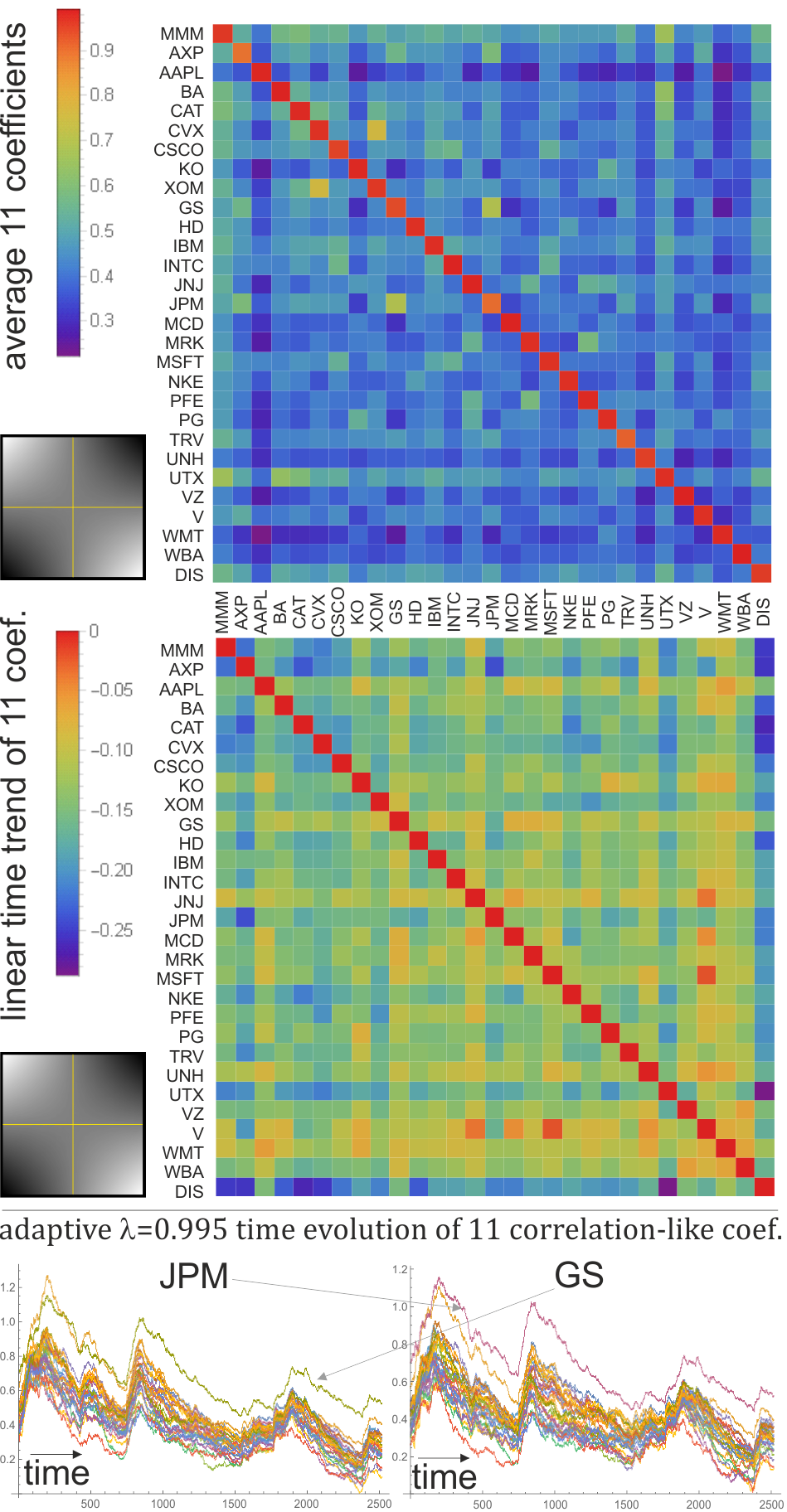}
        \caption{Top: pairwise joint distribution for pairs of normalized $x$ variables would be nearly uniform on $[0,1]^2$ if uncorrelated - the presented matrices contain coefficients for ten-year average of $f_1(x_{i_\alpha})f_1(x_{i_\beta})$ correction to this uniform joint distribution for all pairs (upper) and their linear in time coefficient (lower): $f_1(x_\alpha)f_1(x_\beta)f_1(t/n)$ - linear trend over this time period, obtained by using time normalized to $[0,1]$ as additional variable like in the adaptivity section. Average coefficients are nearly identical as for correlation matrix in Fig. \ref{stock} - what supports their cumulant-like interpretation. Linear trends are a new information and surprisingly turn out always negative here - we can see a general trend of companies loosing dependencies with others, probably due to relaxation after 2008 crisis. Bottom: conformation of this general trend of lowering dependencies by its adaptive calculation for 2 chosen companies with all 28 remaining.  }
       \label{coef1}
\end{figure}
\begin{figure}[b!]
    \centering
        \includegraphics{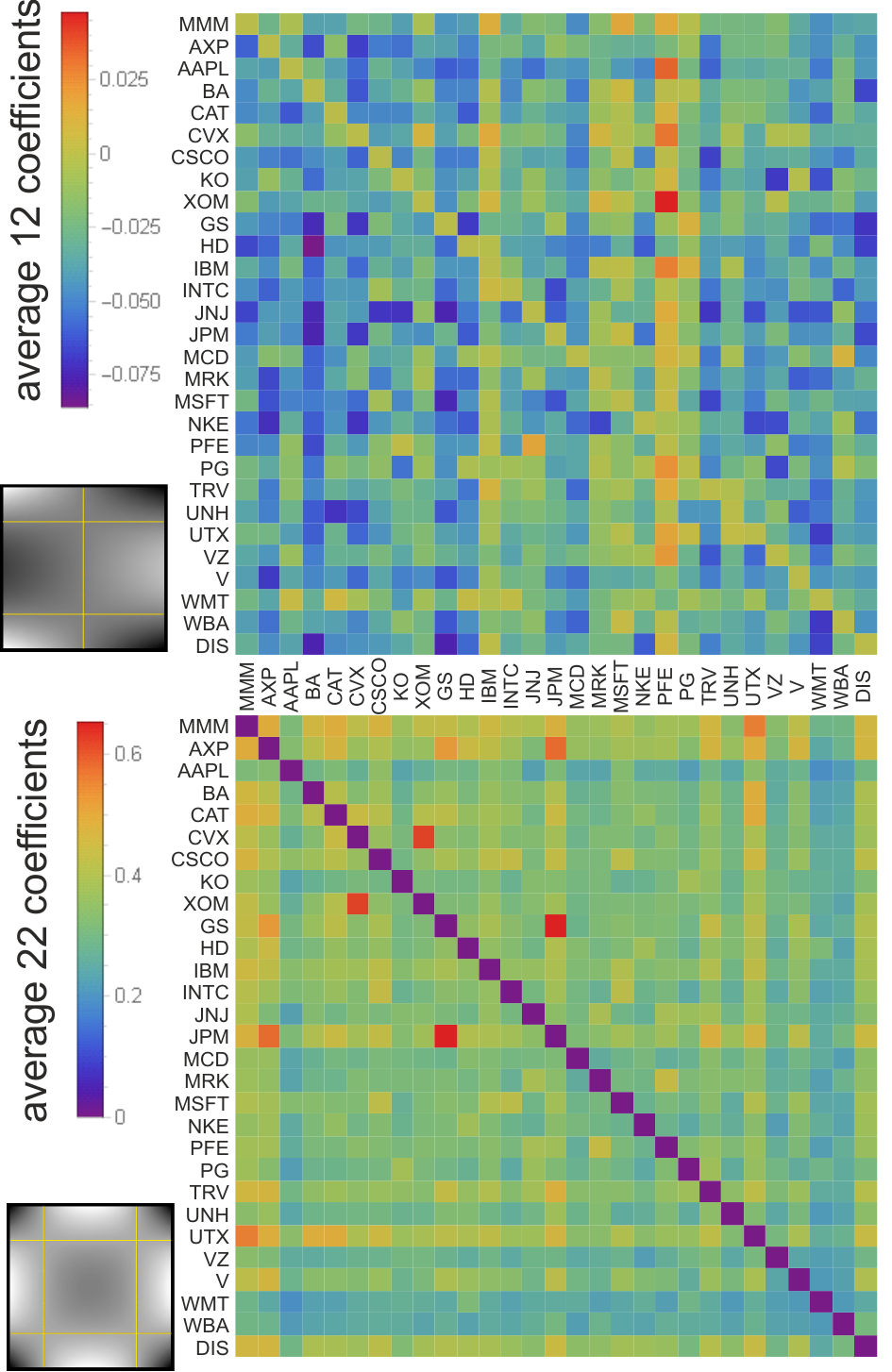}
        \caption{Ten-year average coefficients for two succeeding corrections to uniform pair-wise joint distributions. Top: $f_1(x_\alpha)f_2(x_\beta)$ coefficients describing growth or reduction of variance of the second variable with growth of the first variable. Obtained matrix is no longer symmetric, we can clearly see some columns of companies having different influence on variance of others. We could also find its linear time trend as coefficients of $f_1(x_\alpha)f_2(x_\beta)f_1(t/n)$. Bottom: $f_2(x_\alpha)f_2(x_\beta)$ coefficients describing dependencies of variance - it turns out always positive here, corresponding to coappearance of extreme values - dark edges in Fig. \ref{cor}. Its linear time trend is not presented, but turned out always negative like for $f_1(x_\alpha)f_1(x_\beta)$, meaning weakening of dependencies. Diagonal terms are meaningless so they are filled with 0. While these corrections are formally degree 3 or 4 polynomials, restricting to only some of them allows to use eigenvectors of above matrices.   }
       \label{coef2}
\end{figure}

The standard approach to quantitatively describe statistical dependencies between variables is by correlation matrix - presented in Fig. \ref{stock}. We can use discussed here cumulant-like coefficients for better undersetting of more complex statistical dependencies and their time trends, each of them having unique specific interpretation.

The simplest application is modelling pairwise statistical dependencies - this time spatial instead of temporal as before. After normalization of all variables to nearly uniform distribution (with Laplace CDF like previously), pairwise joint distribution would be nearly uniform on $[0,1]^2$ if they are uncorrelated. To this $\rho_\emptyset=1$ initial density we add independently calculated correction as products of two orthonormal polynomials - coefficients of the first three for all pairs are presented in Fig. \ref{coef1} and \ref{coef2}. There are also presented linear time trends for two of them: using first coefficient with time rescaled to $[0,1]$ as additional variable like in the previous section. The diagonal terms have no meaning in this methodology - are filled with a constant value.

These additional coefficients allow for deeper understanding of statistical dependencies, like between growth of one variable and variance of the second (12) or between their variances (22) like in ARCH. Time trends (calculated for previous 10 years) may suggest direction of further evolution of these dependencies. These coefficients are similar to multivariate mixed cumulants, but having a direct translation to probability density.

Figure \ref{stock} also contains first 5 eigenvectors of covariance matrix as in standard PCA technique. They correspond to largest variance directions and often turn out to group dependent variables. Here we usually estimate density as higher degree polynomials (formally $\sum_i j_i$), for which there are generalizations of eigenvector decomposition~\cite{hyper}, but they are more difficult to calculate and interpret. Presented matrices use only subset of coefficients, allowing to use standard eigenvector decomposition, e.g. $[c_{\alpha\beta}]v^k=\lambda_k v^k$:
$$\sum_{\alpha\beta} c_{\alpha\beta} f_2(x_\alpha)f_2(x_\beta) = \sum_{k=1}^n \lambda_k \left(\sum_\alpha v_\alpha^k f_2(x_\alpha)\right)^2 $$
Presented matrices describe only pair-wise statistical dependencies, we can analogously model dependencies in triples e.g. $f_1(x_\alpha)f_1(x_\beta)f_1(x_\gamma)$ or in larger groups, and their time evolutions.

Log-likelihood evaluation and examples of modelling pairwise joint distributions are presented in Fig. \ref{eval}. We can see how problematic is being fixed on using only Gaussian distribution, what leaves a burden of far inferior predictions - in terms of both unconditional coverage (Fig. \ref{cor}) and log-likelihood. If choosing base distribution accordingly to data (like Laplace), exploiting further statistical dependencies (3 diagrams) leaves not much place for further improvements (from $\approx 1.03$ to $1.09$ times larger average density) for this type of data. Generally possible improvement is bounded by conditional entropy: average over possible contexts of entropy of the new value, which asymptotically is minus average log-likelihood. For example if we would like to encode these values with $1/2^q$ accuracy, for large $q$ we would need on average $\approx q$ minus log-likelihood bits per value.

The discussed here approach starts with normalizing all variables to nearly uniform distribution. Alternative approach is to multiply idealized distribution (e.g. multivariate Gaussian from PCA, or Laplace) by polynomial, estimating the coefficients~\cite{me1}.

\section{Extensions}
The used examples presented basic methodologies for educative reasons, which in real models can be optimized, for example:
\begin{itemize}
  \item Selective choice of basis: we have used complete basis of polynomials, what makes its $(m+1)^d$ size impractically large especially for high dimensions. However, usually only a small percentage of coefficients is above noise - we can selectively choose and use a sparse basis of significant values instead - describing real statistical dependencies, mixed moments between only a small number of variables. Another option is to selectively reduce polynomial degree for some of variables, or for example restrict the real degree of the entire polynomial: $\sum_i j_i$ instead of each coordinate.
  \item Long-range value prediction: combining with state-of-art prediction models exploiting long-range dependencies, for example using a more sophisticated (than just the previous value) predictor of the current value.
  \item Improving informational content of context used for prediction: instead of using a few previous values as the context, we can use some features e.g. describing long-range behavior like average over a time window, or for example obtained from dimensionality reduction methods like PCA (principal component analysis).
  \item Maybe directly combining with a successful e.g. ARMA/ARCH-like model: use it for normalization $x^t=\textrm{CDF}_t (y^t)$ using its modeled distributions like ARCH Gaussian of varying width, then model e.g. $(x^{t-1},x^t)$ with polynomial to extract and exploit further statistical dependencies from data sample.
  \item Multivariate time series usually allow for much better prediction, as presented in \cite{me4}. Adding for example macroeconomic variables should improve prediction. \end{itemize}

\appendix
This appendix contains Wolfram Mathematica source for basic discussed procedure and stationary process, optimized to use built-in vector operations:
\begin{scriptsize}
\begin{verbatim}
im = Import["c:/djia-100.xls"];
v = Log[Transpose[im[[1]]][[2, 2 ;; -1]]];
Print[ListPlot[v]];
n0 = Length[v];
yt = Table[v[[i + 1]] - v[[i]], {i, n1 = n0 - 1}];
syt = Sort[yt];                (* for approximated CDF *)
mu = Median[yt];                 (* Laplace estimation *)
b = Mean[Abs[yt - mu]];
cdfL = If[y < mu, Exp[(y-mu)/b]/2, 1-Exp[-(y-mu)/b]/2];
Print["Laplace distribution: mu= ", mu, "  b= ", b];
Print[Show[
   ListPlot[Table[{syt[[i]], (i - 0.5)/n1}, {i, n1}]],
    Plot[cdfL, {y, -0.1,0.1},PlotStyle -> {Thin, Red}]]];
xt = Table[cdfL /. y -> yt[[i]],{i,n1}]; (* normalized *)
Print[ListPlot[Sort[xt]]]; Print[ListPlot[xt]];
cl = 3; d = 1 + cl;  (* dimension = 1 + context length *)
m = 4;                 (* maximal degree of polynomial *)
coefn = Power[m + 1, d]; Print[coefn, " coefficients"];
p = Table[Power[x, k], {k, 0, m}];
p = Simplify[Orthogonalize[p,Integrate[#1 #2,{x,0,1}]&]];
Print["used orthonormal polynomials: ", p];
n = n1 - cl;            (* final number of data points *)
(* table of contexts and their polynomials: *)
ct = Transpose[Table[xt[[i + cl ;; i ;; -1]], {i, n}]];
ctp = Table[
   If[j==1, Power[ct,0], p[[j]] /. x -> ct], {j, m+1}];
(* calculate coefficients: *)
coef = Table[jt = IntegerDigits[jn, m + 1, d] + 1;
       Mean[Product[ctp[[jt[[c]], c]], {c, d}]],
       {jn, 0, coefn - 1}];

(* find 1D polynomials for various times: *)
pt = Table[0, {i, m + 1}, {i, n}];
Do[jt = IntegerDigits[jn, m + 1, d] + 1;
  pt[[jt[[1]]]] +=
  coef[[jn+1]] * Product[ctp[[jt[[c]], c]],
  {c, 2, cl + 1}], {jn, 0, coefn - 1}];
(* probability normalization to 1: *)
Do[pt[[i]] /= pt[[1]], {i, m + 1, 1, -1}];
(* predicted densities for observed values: *)
rho = Sum[ctp[[i, 1]] * pt[[i]], {i, m + 1}];
Print[ListPlot[Sort[rho]]];
(* densities in 10 random times: *)
plst = RandomInteger[{1, n}, 10];
pl = Table[i = plst[[k]];
  Sum[pt[[j, i]]*p[[j]], {j, m + 1}], {k, Length[plst]}];
Plot[pl, {x, 0, 1}, PlotRange -> {{0, 1}, {0, 5}}]
\end{verbatim}
\end{scriptsize}

\bibliographystyle{IEEEtran}
\bibliography{cites}
\end{document}